\def\mathnlpappendix{1}
\documentclass[11pt]{article}

\usepackage[preprint]{acl}
\usepackage{times}
\usepackage{latexsym}
\usepackage[T1]{fontenc}
\usepackage[utf8]{inputenc}
\usepackage{microtype}
\usepackage{inconsolata}
\usepackage{amsmath}
\usepackage{booktabs}
\usepackage{graphicx}
\usepackage{fancyhdr}

\title{Predicting the Next State Is Not Enough:\\JEPA Representations for Lean Theorem Proving}
\author{
Aarnav Choudhary \\
University of California, Los Angeles \\
\texttt{aarnav11@g.ucla.edu}
}

\newcommand{\state}{s}
\newcommand{\tactic}{a}
\newcommand{\nextstate}{s'}
\newcommand{\appendixref}[1]{%
  \ifdefined\mathnlpappendix Appendix~\ref{#1}\else the supplementary material\fi}

\begin{document}
\maketitle
\thispagestyle{fancy}

\begin{abstract}
Neural theorem provers must both propose tactics and decide which valid
successor states to explore. We study whether one-step Lean transitions
provide a self-supervised signal for branch ordering.  A JEPA-style model
predicts latent successor representations and
scores only kernel-validated, nonterminal successors generated by a fixed
pretrained ByT5 proposer.  JEPA achieves higher Top-1 than matched InfoNCE on a
same-theorem ranking diagnostic (50.18\% versus 31.55\%), but averages 282.3 of
987 solved theorems across three seeds versus 308 for proposer ordering, while
requiring more tactic checks.  In this setting, accurate one-step transition
ranking is therefore insufficient as a long-horizon search value.  The
controlled evaluation separates representation from proposal quality and
treats kernel-checked proof completion as the primary endpoint.
\end{abstract}

\section{Introduction}

Lean proof search alternates between proposing a tactic and asking a trusted
kernel to validate its effect on the current state \citep{demoura2015lean}.
Although modern systems learn the proposal distribution, success also depends
on which valid branches receive the remaining search budget
\citep{yang2023leandojo,george2025leanprogress,lamont2025threedprover}.
We ask whether a one-step proof transition can provide a self-supervised
branch-ordering signal without proof-length or success labels.

Proof artifacts have supported self-supervised auxiliary tasks for tactic
generation \citep{han2022pact}.  We instead adapt joint-embedding predictive
architectures (JEPAs), which predict target representations rather than target
tokens \citep{assran2023ijepa}, to transitions with context
$(\state,\tactic)$ and recorded successor $\nextstate$.  We use an existing
pretrained ByT5 tactic generator \citep{xue2022byt5,yang2023leandojo}, keep its
parameters and deterministic decoding fixed, and use its candidates in every
condition; the transition model does not generate tactics.  Because proof
states and tactics form a
structured mathematical language, evaluating representations learned from them
is relevant to automated theorem proving, autoformalization, and
language-based mathematical reasoning whose outputs require kernel verification.

Our contributions are: (i) a self-supervised latent proof-transition objective
with an exponential-moving-average (EMA) target encoder; (ii) a real-kernel
beam-search integration in which learned scores never see an unvalidated
successor; and (iii) a leakage-controlled comparison with proposer score,
hand-designed progress, and a matched InfoNCE objective.  Both diagnostic and
end-to-end outcomes are reported under the same fixed experimental protocol.

The evaluation separates three claims: distinguishing a recorded successor
from states of other theorems, retaining that signal against harder
same-theorem alternatives, and improving allocation of a finite kernel-search
budget.  The first two act as representation diagnostics, only the third
measures actual performance.

\begin{figure*}[t]
  \centering
  \includegraphics[width=\textwidth]{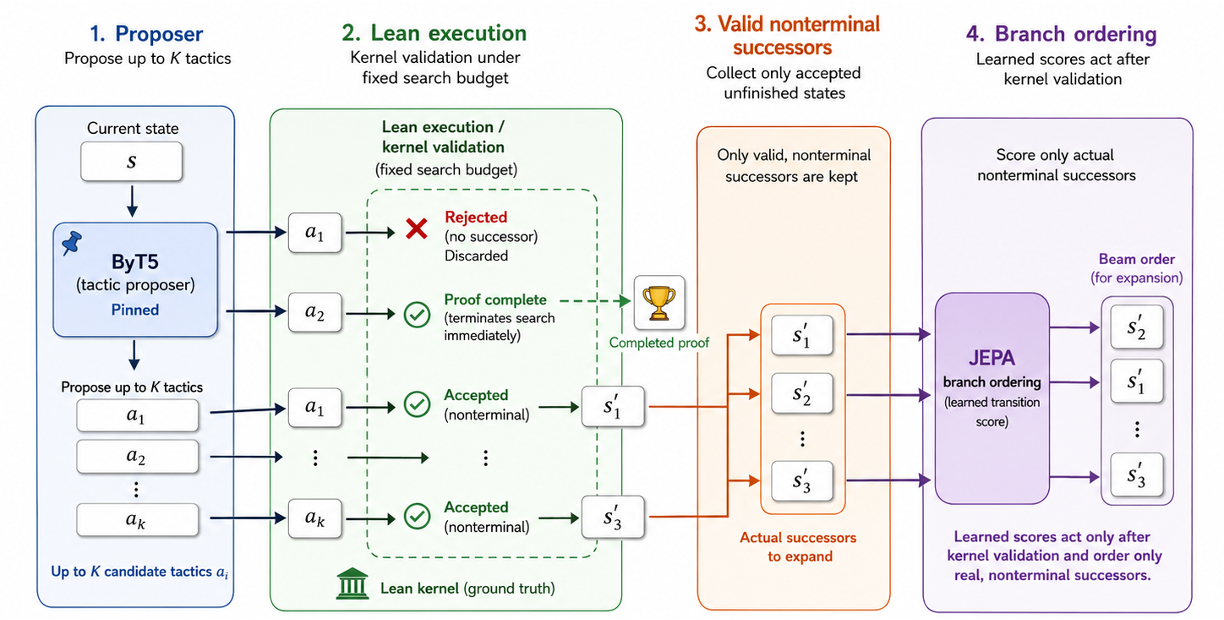}
  \caption{The search pipeline: a pinned ByT5 model proposes up to \textit{K} tactics, which are executed and validated by Lean; invalid tactics are discarded and completed proofs end the search. The JEPA model scores only valid nonterminal successor states, influencing branch expansion order without affecting tactic generation or correctness.}
  \label{fig:overview}
\end{figure*}

\section{Method}

\subsection{Self-Supervised Transition Objective}

For a recorded one-step transition $(\state,\tactic,\nextstate)$, a context encoder
$f_\theta$, predictor $q_\phi$, and target encoder $f_\xi$ produce
\begin{equation}
  \hat{z}=q_\phi(f_\theta([\state;\tactic])), \qquad
  z'=\operatorname{sg}(f_\xi(\nextstate)),
\end{equation}
where $\operatorname{sg}$ stops gradients and
$\xi\leftarrow m\xi+(1-m)\theta$.  Our JEPA loss is
\begin{equation}
  \mathcal{L}_{\mathrm{J}}
  =1-\cos(\hat{z},z')
   +\lambda_v\mathcal{L}_{\mathrm{var}}(\hat{z})
   +\lambda_c\mathcal{L}_{\mathrm{NCE}}(\hat{z},z').
\end{equation}
We use JEPA for this predictive objective with a low-weight contrastive
auxiliary and InfoNCE for the matched direct-contrastive baseline.

Following the variance regularizer proposed by \citet{bardes2022vicreg}, the
variance floor discourages collapse.  InfoNCE uses
$\mathcal{L}_{\mathrm{NCE}}$ directly; encoders, predictor, EMA, data,
optimizer, batch, epochs, and seeds are identical.

\subsection{Kernel-Gated Beam Search}

At each node, ByT5 proposes up to $K$ tactics; Lean rejects invalid tactics
before scoring and terminates immediately on a completed proof.  Each valid
nonterminal successor receives a step score---the proposer's sequence score,
$-g(s')-10^{-6}|s'|$ for progress ($g$ counts goals), or predicted--target
cosine---and its beam priority is the sum of step scores along its partial
proof.  Because scores are not cross-calibrated, comparisons concern each
method's induced ordering.  We keep the higher-priority path for duplicate printed states, sort
descending, retain the top eight, and preserve proposal traversal order on
exact ties.

Figure~\ref{fig:overview} illustrates that JEPA scores only kernel-validated, nonterminal successors and therefore cannot reduce the kernel cost already spent on candidate execution; it only allocates the remaining budget among valid, unfinished branches.

Unlike LeanProgress, which uses supervised remaining-step targets
\citep{george2025leanprogress}, our objective requires only recorded one-step
transitions.  3D-Prover predicts multiple tactic properties and performs
diversity-aware tactic selection \citep{lamont2025threedprover}, whereas our
model orders kernel-produced successor states.  Effect-grounded tactic
embeddings are evaluated through retrieval or analogy
\citep{samoylov2025operators}; our primary endpoint is completed,
kernel-checked proofs.

\section{Experimental Setup}

\paragraph{Data integrity.}
LeanDojo Benchmark 4 contains 259,580 transitions in the official random split
\citep{yang2023leandojo}: 250,814 training (96.62\%), 4,260 validation (1.64\%),
and 4,506 test (1.74\%).  They span 59,557 training, 995 validation, and 992
test theorems, with zero overlap.  We use the complete training and test
partitions.  Imported supervised fine-tuning (SFT) rows match every
state pair and identity in the official traced archive.  They are
treated as independent transitions: only 67.3\% of adjacent within-theorem
pairs are state-continuous because nested goals do not form a linear trace.
The byte-BPE tokenizer is fit on train text only.  Empty transition fields and
cross-split overlap are zero; the artifact file and every checkpoint are
checksummed.  Either the state--tactic input or successor exceeds 1,024 tokens
in 0.38\%, 0.35\%, and 0.31\% of training, validation, and test rows,
respectively.

\paragraph{Models and training.}
The JEPA-style model and direct InfoNCE control both use six Transformer
layers, width 512, a 384-dimensional
projection, and a three-layer predictor: 22.7M trainable parameters plus a
21.5M-parameter EMA target.  Maximum length is 1,024 tokens.  We train ten
epochs at physical batch 32 and gradient accumulation 16 (effective batch 512)
for seeds 7, 17, and 37.  JEPA uses $m=0.996$, $\lambda_v=0.1$, and
$\lambda_c=0.1$.  Architecture, optimization, EMA, data, checkpoint selection,
and seeds are matched between the two scorers; only the objective differs.
InfoNCE uses the
other 31 physical-batch examples as negatives; accumulation changes the
optimizer batch, not that negative pool.
The untrained control instantiates the same architecture and train-fitted
tokenizer independently at each seed, without copying learned weights.
The supervised control is the fixed published tactic generator's sequence
likelihood.  We do not fabricate a matched remaining-step target: the imports
lack verified distance labels, so deterministic goal-count/state-length
progress is the separate value heuristic.

Proposer score preserves the tactic generator’s learned prior; progress provides a hand-designed value heuristic; the untrained model isolates effects of the architecture and tokenizer without representation learning; and direct InfoNCE tests whether predictive learning helps beyond contrastive discrimination.

\paragraph{Ranking evaluation.}
Cross-theorem ranking uses all 4,506 test transitions with 16, 32, and 64
negatives from other theorems.  Same-theorem ranking uses 1,290 transitions
having at least 16 distinct alternatives and evaluates 1, 4, and 16 negatives.
A fixed seed gives every checkpoint identical pools within each scope.  We
report Top-1, MRR, true/negative cosine, and margin.
\paragraph{Search configuration.}
Search uses the existing pretrained LeanDojo ByT5-small tactic generator at
pinned revision \texttt{67a2c53} \citep{yang2023leandojo}, with frozen
parameters and deterministic decoding, LeanDojo 2.1.3, and Lean 4.10.0-rc1.
Conditions share theorem keys and the deterministic proposer configuration;
learned conditions are run for all three seeds.
Completed terms containing \texttt{sorry} or unresolved
metavariables are rejected before a final kernel type check. The frozen
budget is $K=64$, 512 tactic checks, beam 8,
depth 16, and 120 seconds per theorem. Each condition covers all 992 test keys
exactly once, with the theorem set partitioned into five deterministic stride
shards, each fixed to one RTX 3090 device across two hosts for every condition.
Root initialization retains LeanDojo's 600-second default; tactics retain a
60-second limit.
\paragraph{Failure handling.}
A tactic-level REPL exit consumes its check and triggers
at most 10 exact-state recoveries.
Exhausting the restart bound invalidates the entire condition.
Deterministic zero-check backend-initialization failures are listed and
excluded uniformly rather than counted as failed proofs; we report both 992
attempted keys and the eligible denominator. We report theorem-level wins/losses
and a paired bootstrap 95\% interval for solve-rate differences.  Any missing
key or budget mismatch invalidates a condition.  An unrecoverable in-search
kernel-process error terminates that theorem as an unsolved failure and is
reported per condition rather than excluded.
We predefine material check efficiency as identical solved count in every seed,
at least 10\% fewer ordinary checks in every seed, a theorem-clustered check
interval below zero, and no increase in theorem timeouts.
We also report valid/checked tactic applications (\texttt{step\_success}) and
termination/failure counts.
\paragraph{Statistical analysis.}
The theorem, rather than a tactic application or training seed, is the unit of
paired inference.  For learned conditions we first average the three outcomes
within each theorem and then resample theorem keys, retaining their matched
outcomes across search orders.  This avoids treating correlated seeds or the
many checks inside one proof attempt as independent evidence.  The primary
claim is fixed before inspecting test outcomes. A positive result requires
more solved theorems, or exactly matched solves with the preregistered check
reduction, however ranking metrics alone cannot satisfy it.

\begin{table}[t]
  \centering
  \scriptsize
  \setlength{\tabcolsep}{2.2pt}
  \begin{tabular}{lrrrr}
    \toprule
    & \multicolumn{2}{c}{Same theorem (16)} & \multicolumn{2}{c}{Cross theorem (64)} \\
    \cmidrule(lr){2-3}\cmidrule(lr){4-5}
    Objective & Top-1 & MRR & Top-1 & MRR \\
    \midrule
    Untrained & 0.0708 & 0.1985 & 0.0200 & 0.0825 \\
    JEPA & \textbf{0.5018} & \textbf{0.6423} & 0.7837 & 0.8057 \\
    InfoNCE & 0.3155 & 0.4754 & \textbf{0.8105} & \textbf{0.8422} \\
    \bottomrule
  \end{tabular}
  \caption{Test transition ranking, reported as three-seed means.  Bold marks
  the best objective within each negative scope; full uncertainties and cosine
  margins appear in \appendixref{app:ranking}.}
  \label{tab:ranking}
\end{table}

\section{Results and Discussion}

Table~\ref{tab:ranking} reports the preregistered representation diagnostics.
On the harder same-theorem population of 1,290 transitions, JEPA achieves
$0.5018{\pm}0.0072$ Top-1, compared with $0.3155{\pm}0.0074$ for InfoNCE and
$0.0708{\pm}0.0256$ for untrained weights, and leads both controls in every
seed.  Holding theorem identity constant removes an easy retrieval shortcut,
although the task remains action-conditioned and does not measure proof
completion.

With 64 cross-theorem negatives, InfoNCE instead achieves the highest Top-1
($0.8105{\pm}0.0072$), followed by JEPA ($0.7837{\pm}0.0020$).  The reversal
shows that objective rankings depend on negative construction; neither
diagnostic alone establishes a useful long-horizon search ordering.

\begin{table}[t]
  \centering
  \footnotesize
  \setlength{\tabcolsep}{1.9pt}
  \begin{tabular}{lrrr}
    \toprule
    Search order & Solved & Checks & Wall (s) \\
    \midrule
    Proposer score & \textbf{308/987} & \textbf{349.92} & \textbf{64.62} \\
    Progress & 304/987 & 352.99 & 65.31 \\
    Untrained & $303.7{\pm}8.5$/987 & $351.03{\pm}2.86$ & $65.99{\pm}0.30$ \\
    InfoNCE & $289.7{\pm}3.1$/987 & $359.89{\pm}0.50$ & $66.02{\pm}0.03$ \\
    JEPA & $282.3{\pm}1.5$/987 & $363.45{\pm}1.02$ & $66.61{\pm}0.08$ \\
    \bottomrule
  \end{tabular}
  \caption{Fixed-budget Lean-kernel search.  Checkpoint rows report three-seed
  means and sample standard deviations.  Bold marks the best value: higher for
  solved theorems and lower for checks and wall time.  Paired comparisons use
  theorem-level outcomes.}
  \label{tab:search}
\end{table}

Table~\ref{tab:search} gives the primary endpoint.  JEPA solves
$282.3{\pm}1.5$ of 987 eligible theorems, compared with 308 for proposer
ordering.  The theorem-clustered mean difference is $-25.67$ solves
($-2.600$ percentage points; 95\% interval $[-3.715,-1.554]$), accompanied by
13.53 additional checks per theorem (95\% interval $[9.48,17.85]$).  JEPA also
trails progress, untrained weights, and InfoNCE; complete paired comparisons
appear in \appendixref{tab:jepa-paired}.  One in-search backend error in each
of InfoNCE seed 7 and JEPA seed 17 is conservatively counted as unsolved, with
no condition-specific exclusion.

The preregistered positive gate is not satisfied.  Under this proposer and
budget, accurate ranking of observed local transitions does not produce a
useful ordering for proof completion.  This result is consistent with an
objective mismatch: transition consistency characterizes the immediate effect
of a tactic, whereas search requires the long-horizon value of the resulting
branch.

The matched untrained model stays close to proposer and progress, while both
trained representations solve fewer theorems.  Because untrained, JEPA, and
InfoNCE share scoring cost and architecture, inference overhead alone is
insufficient to explain the learned models' larger solve deficits.  The data
do not identify a unique mechanism, but they localize the failure to the
ordering induced by the learned objectives under this candidate distribution.
In particular, a locally predictable successor may preserve the surface form
and tactic effect of its parent without being a strategically valuable state
from which to finish the theorem.

These findings indicate that held-out transition retrieval should be treated
as a representation diagnostic rather than a prover result.  Kernel-aligned
evaluation must additionally fix the candidate source, charge every tactic
execution to a common budget, and report completed proofs and checks.
Longer-horizon or set-valued JEPA targets remain hypotheses requiring separate
preregistered evaluation.

\section{Conclusion}

We evaluate JEPA-style proof-transition learning as an ordering mechanism for
successors produced by Lean.  Fixing the generator, theorem set, and
kernel-check budget separates representation quality from proposal quality and
replay artifacts.  Across three seeds, JEPA provides the strongest
same-theorem transition ranking but solves fewer proofs and consumes more
checks than every control.  The study therefore yields a controlled negative
result under this configuration for one-step latent prediction as a
branch-ordering score, without addressing longer-horizon JEPA objectives.

\section*{Limitations}

Our results are limited to one small encoder, one tactic generator, and one
random split.  The SFT rows contain independent one-step transitions rather
than verified trajectories, precluding trajectory objectives and a learned
remaining-step control.  Successor scoring follows kernel execution, long
states are truncated at the reported rates, and wall-clock measurements are
hardware-dependent; the tactic-check budget and paired solved set are therefore
the primary controls.  Training and search also differ in transition
distribution: recorded one-step SFT examples supervise the model, whereas
search states are induced by the fixed ByT5 candidate set.  The design does
not distinguish objective mismatch from out-of-distribution scoring.  In
addition, each ordering accumulates its native step score, so depth preferences
can differ even though the aggregation rule is fixed before test evaluation.
Random-split library similarity may also limit generalization; all require
separate evaluation.
\section*{Acknowledgements}

I would like to thank Benjamin Yu (UCLA) and Leon Lenk for their advice, helpful discussions, feedback, and support throughout this work. It would not have been possible without them.

\bibliography{references}

\ifdefined\mathnlpappendix
\clearpage
\appendix
\definecolor{appBlue}{HTML}{0072B2}
\definecolor{appOrange}{HTML}{E69F00}
\definecolor{appGreen}{HTML}{009E73}
\definecolor{appPurple}{HTML}{CC79A7}
\definecolor{appGray}{HTML}{666666}

\newcommand{\appnode}[3]{%
  \fcolorbox{#1}{#1!8}{%
    \parbox[c]{#2}{\centering\strut #3\strut}}}

\section{Evidence Chain}
\label{app:roadmap}

The study follows one question: does learning a valid one-step transition
signal improve the order in which a prover spends its search budget?  The
appendix follows that question from representation learning to the
kernel-checked endpoint.

\begin{figure*}[t]
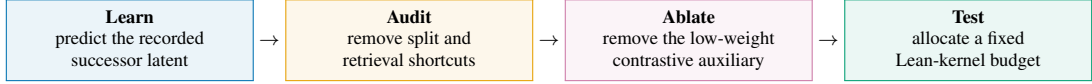

  \centering
  \scriptsize
  \setlength{\tabcolsep}{2pt}
  \begin{tabular}{@{}c@{\;$\rightarrow$\;}c@{\;$\rightarrow$\;}c@{\;$\rightarrow$\;}c@{}}
    \appnode{appBlue}{.19\textwidth}{\textbf{Learn}\\predict the recorded\\successor latent}
    & \appnode{appOrange}{.19\textwidth}{\textbf{Audit}\\remove split and\\retrieval shortcuts}
    & \appnode{appPurple}{.19\textwidth}{\textbf{Ablate}\\remove the low-weight\\contrastive auxiliary}
    & \appnode{appGreen}{.19\textwidth}{\textbf{Test}\\allocate a fixed\\Lean-kernel budget} \\
  \end{tabular}
  \caption{The evidence chain.  Local ranking establishes representation
  quality; only completed, kernel-checked proofs establish search value.}
  \label{fig:evidence-chain}
\end{figure*}

\section{Model and Evaluation Boundary}
\label{app:architecture}

\begin{figure*}[t]
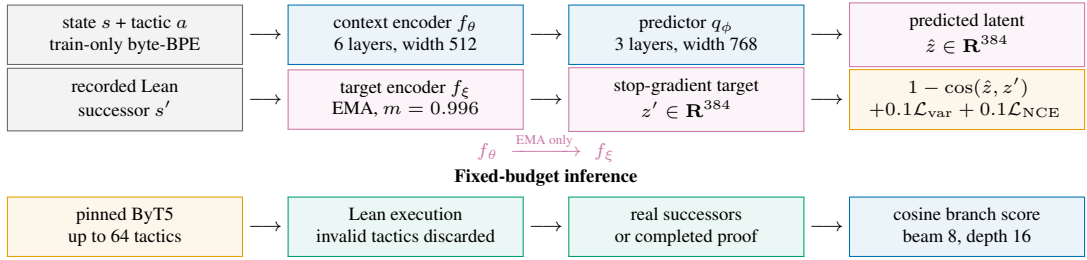

  \centering
  \scriptsize
  \setlength{\tabcolsep}{2pt}
  \begin{tabular}{@{}c@{\;$\longrightarrow$\;}c@{\;$\longrightarrow$\;}c@{\;$\longrightarrow$\;}c@{}}
    \appnode{appGray}{.18\textwidth}{state $s$ + tactic $a$\\train-only byte-BPE}
      & \appnode{appBlue}{.18\textwidth}{context encoder $f_\theta$\\6 layers, width 512}
      & \appnode{appBlue}{.18\textwidth}{predictor $q_\phi$\\3 layers, width 768}
      & \appnode{appPurple}{.18\textwidth}{predicted latent\\$\hat z\in\mathbf{R}^{384}$} \\
    \appnode{appGray}{.18\textwidth}{recorded Lean\\successor $s'$}
      & \appnode{appPurple}{.18\textwidth}{target encoder $f_\xi$\\EMA, $m=0.996$}
      & \appnode{appPurple}{.18\textwidth}{stop-gradient target\\$z'\in\mathbf{R}^{384}$}
      & \appnode{appOrange}{.18\textwidth}{$1-\cos(\hat z,z')$\\$+0.1\mathcal L_{\rm var}+0.1\mathcal L_{\rm NCE}$} \\
    \multicolumn{4}{c}{\vspace{2pt}\textcolor{appPurple}{$f_\theta\;\xrightarrow{\text{EMA only}}\;f_\xi$}} \\
    \multicolumn{4}{c}{\vspace{3pt}\textbf{Fixed-budget inference}} \\
    \appnode{appOrange}{.18\textwidth}{pinned ByT5\\up to 64 tactics}
      & \appnode{appGreen}{.18\textwidth}{Lean execution\\invalid tactics discarded}
      & \appnode{appGreen}{.18\textwidth}{real successors\\or completed proof}
      & \appnode{appBlue}{.18\textwidth}{cosine branch score\\beam 8, depth 16} \\
  \end{tabular}
  \caption{Training and inference boundary.  The learned model predicts and
  orders representations; Lean alone creates successors and certifies proof
  completion.}
  \label{fig:architecture}
\end{figure*}

The target encoder receives no optimizer gradient.  During search, JEPA scores
only nonterminal successors returned by Lean; invalid tactics are already
discarded and completed proofs already terminate.  Thus scoring can reallocate
the remaining budget, but cannot recover the check spent producing a
successor.

\begin{table*}[t]
  \centering
  \footnotesize
  \setlength{\tabcolsep}{4pt}
  \begin{tabular}{ll@{\qquad}ll}
    \toprule
    Setting & Value & Setting & Value \\
    \midrule
    Tokenizer & byte-BPE, train only & Maximum length & 1,024 \\
    Vocabulary / min. frequency & 16,000 / 2 & Encoder layers / heads & 6 / 8 \\
    Hidden / feed-forward width & 512 / 1,024 & Projection / predictor width & 384 / 768 \\
    Predictor layers / dropout & 3 / 0.15 & Trainable / EMA parameters & 22.7M / 21.5M \\
    Epochs / seeds & 10 / 7, 17, 37 & Physical / effective batch & 32 / 512 \\
    Learning rate / weight decay & $10^{-5}$ / $10^{-4}$ & Gradient clip / accumulation & 1.0 / 16 \\
    EMA / InfoNCE temperature & 0.996 / 0.05 & Variance / auxiliary weight & 0.1 / 0.1 \\
    \bottomrule
  \end{tabular}
  \caption{Matched model and optimization settings.  Direct InfoNCE changes
  the objective; untrained controls use the same architecture and tokenizer.}
  \label{tab:hyperparameters}
\end{table*}

The best validation-loss checkpoint is selected within each fixed ten-epoch
run.  Accumulation changes the optimizer batch, not the 31-example in-batch
negative pool.  Random controls are independently initialized at each seed.

\section{Local Transition Evidence}
\label{app:ranking}

The first large JEPA collapsed: true and negative successor cosines were both
$0.9537$, yielding essentially zero margin.  A variance floor and low-weight
contrastive auxiliary restored separation.  The early rows below explain the
model-development path; only the official three-seed rows support claims.

\begin{table*}[t]
  \centering
  \footnotesize
  \setlength{\tabcolsep}{5pt}
  \begin{tabular}{llrrrr}
    \toprule
    Stage & Objective / split & Negatives & Top-1 & MRR & Margin \\
    \midrule
    Initial large model & cosine, artifact split & 16 & 0.0550 & 0.2027 & $-0.0000$ \\
    Stabilized model & cosine + var. + NCE, artifact split & 16 & 0.8944 & 0.9168 & 0.3054 \\
    Clean JEPA & same loss, theorem split & 16 & 0.8862 & 0.9092 & 0.3620 \\
    Clean InfoNCE & direct contrastive, theorem split & 16 & 0.9220 & 0.9548 & 0.2874 \\
    \bottomrule
  \end{tabular}
  \caption{Development diagnostics before the official full-data matrix.
  Populations differ, so rows are provenance rather than a controlled
  comparison.}
  \label{tab:development-ranking}
\end{table*}

The official audit then fixed the unit of evidence.  All 259,580 imported
state pairs and identities match the traced archive, theorem overlap is zero,
and the tokenizer is trained only on the training split.  Encounter order is
not a proof trajectory: only 67.27\% of adjacent within-theorem pairs are
state-continuous, often because nested goals reappear after a printed
\texttt{no goals}.  Training therefore uses independent one-step transitions,
while inference uses states returned directly by Lean.

\begin{table*}[t]
  \centering
  \footnotesize
  \setlength{\tabcolsep}{8pt}
  \begin{tabular}{lrrr}
    \toprule
    Split & Transitions & Theorem keys & Either side over 1,024 tokens \\
    \midrule
    Train & 250,814 & 59,557 & 0.38\% \\
    Validation & 4,260 & 995 & 0.35\% \\
    Test & 4,506 & 992 & 0.31\% \\
    \bottomrule
  \end{tabular}
  \caption{Official Benchmark 4 transition split.  Cross-split theorem
  overlap, empty fields, and source-pair mismatches are all zero.}
  \label{tab:data-audit}
\end{table*}

\begin{table*}[t]
  \centering
  \scriptsize
  \setlength{\tabcolsep}{1.6pt}
  \begin{tabular}{llrrrrr}
    \toprule
    Negative scope & Objective & Top-1 & MRR & True cosine & Neg. cosine & Margin \\
    \midrule
    Cross-theorem, 64 & Untrained & $0.0200{\pm}0.0045$ & $0.0825{\pm}0.0075$ & $-0.0222{\pm}0.0142$ & $-0.0236{\pm}0.0126$ & $0.0013{\pm}0.0021$ \\
    Cross-theorem, 64 & JEPA & $0.7837{\pm}0.0020$ & $0.8057{\pm}0.0024$ & $0.9350{\pm}0.0095$ & $0.6318{\pm}0.0492$ & $0.3032{\pm}0.0398$ \\
    Cross-theorem, 64 & InfoNCE & $\mathbf{0.8105{\pm}0.0072}$ & $\mathbf{0.8422{\pm}0.0056}$ & $0.4281{\pm}0.0151$ & $0.0870{\pm}0.0211$ & $\mathbf{0.3411{\pm}0.0062}$ \\
    \midrule
    Same-theorem, 16 & Untrained & $0.0708{\pm}0.0256$ & $0.1985{\pm}0.0251$ & $-0.0205{\pm}0.0138$ & $-0.0095{\pm}0.0139$ & $-0.0110{\pm}0.0056$ \\
    Same-theorem, 16 & JEPA & $\mathbf{0.5018{\pm}0.0072}$ & $\mathbf{0.6423{\pm}0.0063}$ & $0.9403{\pm}0.0101$ & $0.8178{\pm}0.0408$ & $\mathbf{0.1225{\pm}0.0307}$ \\
    Same-theorem, 16 & InfoNCE & $0.3155{\pm}0.0074$ & $0.4754{\pm}0.0090$ & $0.4730{\pm}0.0157$ & $0.4282{\pm}0.0188$ & $0.0448{\pm}0.0032$ \\
    \bottomrule
  \end{tabular}
  \caption{Hardest official ranking scopes, reported as mean and sample
  standard deviation over seeds.  Raw cosine levels are objective-dependent;
  rank and within-objective margins are the relevant diagnostics.}
  \label{tab:diagnostic-details}
\end{table*}

Cross-theorem negatives favor InfoNCE, whereas same-theorem negatives favor
JEPA in every seed.  Holding theorem identity fixed removes one retrieval
shortcut, but this remains an action-conditioned local diagnostic rather than
proof completion.

\subsection{Post-Hoc Auxiliary-Loss Ablation}

Removing the auxiliary term isolates latent prediction with the same data,
architecture, optimizer, EMA, training budget, and three seeds.

\begin{table*}[t]
  \centering
  \footnotesize
  \setlength{\tabcolsep}{6pt}
  \begin{tabular}{lrrrr}
    \toprule
    Objective & Cross Top-1@64 & Cross MRR@64 & Same Top-1@16 & Same MRR@16 \\
    \midrule
    JEPA, $\lambda_c=0.1$ & $\mathbf{0.7837{\pm}0.0020}$ & $\mathbf{0.8057{\pm}0.0024}$ & $\mathbf{0.5018{\pm}0.0072}$ & $\mathbf{0.6423{\pm}0.0063}$ \\
    JEPA, $\lambda_c=0$ & $0.3880{\pm}0.1347$ & $0.4810{\pm}0.1289$ & $0.3721{\pm}0.0237$ & $0.5113{\pm}0.0269$ \\
    \bottomrule
  \end{tabular}
  \caption{Requested loss ablation on the predefined representation
  diagnostics.  Bold marks the stronger mean within each column.}
  \label{tab:lambda-c-ablation}
\end{table*}

The drop and larger cross-seed variation show that the low-weight contrastive
term materially stabilizes the reported transition signal.  No completed
kernel-search condition exists for $\lambda_c=0$, so this table supports only
a representation-level conclusion.

\section{Kernel-Checked Search}
\label{app:kernel-results}

The operational test fixes Lean 4.10.0-rc1, LeanDojo 2.1.3, ByT5-small revision
\texttt{67a2c53}, 64 candidates, 512 ordinary checks, beam 8, depth 16, and a
120-second theorem limit.  Completed terms containing \texttt{sorry},
unresolved metavariables, or a failed final declaration check are rejected.
REPL recovery replays the exact path, verifies state equality, and never
enlarges the budget.

Five deterministic zero-check initialization failures are excluded uniformly,
leaving 987 eligible theorems from 992 attempted keys.  In-search backend
errors remain eligible unsolved outcomes.

\begin{table*}[t]
  \centering
  \scriptsize
  \setlength{\tabcolsep}{3.4pt}
  \begin{tabular}{lrrrrrr}
    \toprule
    Search order & Solved & Step success & Checks/thm. & Wall s/thm. & Timeouts & Errors \\
    \midrule
    Proposer score & 308/987 & 26.59\% & 349.92 & \textbf{64.62} & 83 & \textbf{0} \\
    Progress & 304/987 & \textbf{28.26\%} & 352.99 & 65.31 & 89 & \textbf{0} \\
    Untrained seed 7 & 294/987 & 25.90\% & 353.52 & 66.23 & 90 & \textbf{0} \\
    Untrained seed 17 & 307/987 & 26.40\% & 351.65 & 66.08 & 91 & \textbf{0} \\
    Untrained seed 37 & \textbf{310/987} & 26.16\% & \textbf{347.91} & 65.65 & 93 & \textbf{0} \\
    JEPA seed 7 & 284/987 & 25.16\% & 364.61 & 66.52 & 84 & \textbf{0} \\
    JEPA seed 17 & 282/987 & 25.42\% & 362.68 & 66.67 & 89 & 1 \\
    JEPA seed 37 & 281/987 & 24.66\% & 363.05 & 66.65 & 85 & \textbf{0} \\
    InfoNCE seed 7 & 289/987 & 24.88\% & 360.41 & 66.00 & \textbf{81} & 1 \\
    InfoNCE seed 17 & 287/987 & 25.32\% & 359.85 & 66.06 & 85 & \textbf{0} \\
    InfoNCE seed 37 & 293/987 & 25.00\% & 359.41 & 66.02 & 88 & \textbf{0} \\
    \bottomrule
  \end{tabular}
  \caption{Complete fixed-protocol outcomes.  Higher is better for solved and
  step success; lower is better elsewhere.  Errors are retained as unsolved.}
  \label{tab:all-kernel-runs}
\end{table*}

\begin{table*}[t]
  \centering
  \footnotesize
  \setlength{\tabcolsep}{5pt}
  \begin{tabular}{lrrrr}
    \toprule
    Comparator & JEPA $\Delta$ solves & $\Delta$ success (points) & 95\% interval & Wins/losses/ties \\
    \midrule
    Proposer score & $-25.67$ & $-2.600$ & $[-3.715,-1.554]$ & 6/36/945 \\
    Progress & $-21.67$ & $-2.195$ & $[-3.445,-1.047]$ & 13/35/939 \\
    Untrained & $-21.33$ & $-2.161$ & $[-3.073,-1.317]$ & 12/49/926 \\
    InfoNCE & $-7.33$ & $-0.743$ & $[-1.317,-0.236]$ & 6/20/961 \\
    \bottomrule
  \end{tabular}
  \caption{Theorem-clustered paired comparisons.  Negative values favor the
  comparator; learned conditions are averaged within theorem over seeds.}
  \label{tab:jepa-paired}
\end{table*}

\begin{figure*}[t]
  \centering
  \footnotesize
  \begin{tabular}{@{}c@{\qquad}c@{\qquad}c@{}}
    \appnode{appOrange}{.25\textwidth}{\textbf{Cross-theorem ranking}\\
      InfoNCE 81.05\%\\JEPA 78.37\%\\\textit{InfoNCE wins}}
    & \appnode{appBlue}{.25\textwidth}{\textbf{Same-theorem ranking}\\
      JEPA 50.18\%\\InfoNCE 31.55\%\\\textit{JEPA wins}}
    & \appnode{appGreen}{.25\textwidth}{\textbf{Kernel completion}\\
      proposer 31.21\%\\JEPA 28.61\%\\\textit{proposer wins}} \\
  \end{tabular}
  \caption{The proxy reversal.  Either representation objective can win a
  local diagnostic while neither induces the best proof-search order.}
  \label{fig:proxy-reversal}
\end{figure*}

\section{Claim Scope and Reproducibility}
\label{app:reproducibility}

The evidence chain supports a narrow conclusion: one-step latent prediction
learns theorem-local transition structure, but that structure is not a useful
long-horizon branch value under this proposer and budget.  It does not rule out
longer-horizon, set-valued, or search-trained JEPA targets.

\begin{table*}[t]
  \centering
  \footnotesize
  \setlength{\tabcolsep}{4pt}
  \begin{tabular}{lll}
    \toprule
    Gate & Failure condition & Accepted evidence \\
    \midrule
    Data & split overlap, empty state, source mismatch & 259,580/259,580 exact pairs \\
    Ranking & incomplete matrix or negative-pool drift & nine linked checkpoints \\
    Kernel & missing key, budget drift, malformed error & 11 conditions, 992 keys each \\
    Pairing & unmatched theorem set or seed as sample & theorem-clustered comparison \\
    Proof validity & \texttt{sorry}, metavariable, declaration failure & kernel-checked completion \\
    Manuscript & placeholder, identifier, unembedded font, cell drift & exact result handoff \\
    \bottomrule
  \end{tabular}
  \caption{Fail-closed gates connecting imported data, checkpoints, search
  outcomes, and reported claims.}
  \label{tab:claim-gates}
\end{table*}

The complete test suite covers target stop-gradient and EMA-only updates,
official theorem disjointness, exact transition provenance, real
Lean-successor scoring, unsafe-proof rejection, restart accounting, shard
completeness, clustered uncertainty, and manuscript-result linkage.  Frozen
artifact and protocol hashes preserve the full audit trail.

\fi

\end{document}